\documentclass[conference]{IEEEtran}

\usepackage{cite}
\usepackage{amsmath,amssymb,amsfonts}
\usepackage{array}
\usepackage[noend]{algpseudocode}
\usepackage[noend,ruled,vlined,linesnumbered]{algorithm2e}
\usepackage{graphicx}
\usepackage{textcomp}
\usepackage{xcolor}
\usepackage{flushend}
\usepackage{multirow}
\usepackage{booktabs}
\usepackage{tabularx}
\usepackage{ragged2e}
\usepackage{comment}
\usepackage[alpha]{mdpn} 
\usepackage[flushleft]{threeparttable} 
\ifCLASSOPTIONcompsoc
  \usepackage[caption=false,font=normalsize,labelfont=sf,textfont=sf]{subfig}
\else
  \usepackage[caption=false,font=footnotesize]{subfig}
\fi

\newcolumntype{L}{>{\RaggedRight\arraybackslash}X} 

\def\BibTeX{{\rm B\kern-.05em{\sc i\kern-.025em b}\kern-.08em
    T\kern-.1667em\lower.7ex\hbox{E}\kern-.125emX}}
\begin{document}

\title{Controlled Deep Reinforcement Learning for Optimized Slice Placement}
\author{\IEEEauthorblockN{Jose Jurandir Alves Esteves\IEEEauthorrefmark{1}\IEEEauthorrefmark{2}, Amina Boubendir\IEEEauthorrefmark{1}, Fabrice Guillemin\IEEEauthorrefmark{1} and Pierre Sens\IEEEauthorrefmark{2}}
\IEEEauthorblockA{\IEEEauthorrefmark{1}Orange Labs, France} \IEEEauthorrefmark{2}Sorbonne Universit\'e / CNRS / Inria, LIP6, France\\ \{josejurandir.alvesesteves, amina.boubendir, fabrice.guillemin\}@orange.com, pierre.sens@lip6.fr }

\maketitle

\begin{abstract}
We present a hybrid ML-heuristic approach that we name "Heuristically Assisted Deep Reinforcement Learning (HA-DRL)" to solve the problem of Network Slice Placement Optimization. The proposed approach leverages recent works on Deep Reinforcement Learning (DRL) for slice placement and Virtual Network Embedding (VNE) and uses a heuristic function to optimize the exploration of the action space by giving priority to reliable actions indicated by an efficient heuristic algorithm. The evaluation results show that the proposed HA-DRL algorithm can accelerate the learning of an efficient slice placement policy improving slice acceptance ratio when compared with state-of-the-art approaches that are based only on reinforcement learning.
\end{abstract}

\begin{IEEEkeywords}
Network Slicing, Optimization, Automation, Deep Reinforcement Learning, Placement, Large Scale
\end{IEEEkeywords}

\section{Introduction}

Network Slice Placement is a critical research problem in the context of Network Slicing \cite{survey_vfnp}, which can be formulated as a multi-objective  Integer Linear Programming (ILP)  optimization problem. It is  well-known in the literature that such a problem is   $\mathcal{NP}$-hard  \cite{vne_np_hardness} with very long convergence time.
Therefore, heuristic approaches have been developed (an extensive list is given in \cite{cnsm_2020}). From an operational perspective, heuristics are more suitable than ILP as they yield faster placement results.
However, they give sub-optimal solutions.  

To overcome convergence issues, Machine Learning (ML) and Reinforcement Learning (RL) techniques have been introduced as they are more accurate than heuristics \cite{p1}. In practice, ensuring that a Deep Reinforcement Learning (DRL) algorithm converges to an optimal policy is a challenge since the algorithm acts as a self-controlled black box. In addition, such algorithm relies on a large number of hyper-parameters to fine-tune in order to ensure accuracy of the solution exploration and the exploitation of knowledge acquired via training. 

We describe the Heuristically Assisted DRL (HA-DRL) introduced in \cite{HA_DRL_TNSM} to specifically address the challenge of obtaining a better and explainable behavior of DRL agents. The proposed solution is innovative: i) because it is a hybrid approach gathering the strength of heuristics and the flexibility of a DRL, and ii) because of its performance results as it accelerates and stabilizes the convergence of state-of-the-art DRL techniques when applied to Network Slice Placement.
In the rest of the paper, we present the problem statement in Section~\ref{sec:pb}. The main aspects of the proposed HA-DRL approach are introduced in Section~\ref{sec:HA-DRL}. 

Some experiments and results are described in Section~\ref{sec:evaluation}, and concluding remarks are given in Section~\ref{sec:conclusion}.

\section{Network Slice Placement Optimization problem \label{sec:pb}}

The compact formulation for the proposed Network Slice Placement Optimization problem is as follows: 
\begin{itemize}
    \item \textit{Given:} a Network Slice Placement Request (NSPR) graph $G_v = (V, E)$ and a Physical Substrate Network (PSN) graph $G_s = (N, L)$,
    \item \textit{Find:} a mapping $G_v \to  \bar{G}_s =(\bar{N},\bar{L})$, $\bar{N} \subset N$, $\bar{L} \subset L$,
    \item\textit{Subject to:} the VNF CPU requirements $req^{cpu}_v, \forall v \in V$, the VNF RAM requirements $req^{ram}_v, \forall v \in V$, the Virtual Link (VL) bandwidth requirements $req^{bw}_{(\bar{a},\bar{b})}, \forall (\bar{a},\bar{b}) \in E$, the node CPU available capacity $cap^{cpu}_n, \forall n \in N$, the node RAM available capacity $cap^{ram}_n, \forall n \in N$, the physical link bandwidth available capacity $cap^{bw}_{(a,b)}, \forall (a,b) \in L$.
    \item \textit{Optimization Objectives: } 1) maximize the network slice placement request acceptance ratio; 2) minimize the total resource consumption, and 3) maximize load balancing.
\end{itemize}

A complete model description and problem formulation can be found in \cite{cnsm_2020} and \cite{HA_DRL_TNSM}. This is a multi-objective optimization problem with objective function given by:
\begin{align}
  \max_{x,y,z} c_{1} z - c_{2} \sum_{(\bar{a},\bar{b}) \in E}\sum_{(a,b) \in L}y^{(\bar{a},\bar{b})}_{(a,b)}req^{bw}_{(\bar{a},\bar{b})} \nonumber \\  + c_{3} \sum_{n\in N}\sum_{v\in V}x^{v}_{n}\left( \frac{cap^{cpu}_{n}}{M^{cpu}_{n}} + \frac{cap^{ram}_{n}}{M^{ram}_{n}} \right)  \label{obj_function}
\end{align}
for some weight coefficients $c_i>0$, $i=1,2,3$, where $M^{cpu}_{n}$ and $M^{ram}_{n}$ are  available capacities on node $n$ in CPU and RAM, respectively. This objective function is clearly a weighted sum of the three optimization objectives described previously and we use the following decision variables:
\begin{itemize}
    \item $x^{v}_{s} \in \{0,1\}$ for ${v} \in V$ and $n \in N$ is equal to 1 if the VNF $v$ is placed onto node $n$ and 0 otherwise,
    \item $y^{(\bar{a},\bar{b})}_{(a,b)} \in \{0,1\}$ for ${(\bar{a},\bar{b})} \in E$ and $(a,b) \in L$ is equal to 1 if the virtual link $(\bar{a},\bar{b})$ is mapped onto physical link $(a,b)$ and 0 otherwise.
    \item $z  \in \{0,1\}$ is equal to 1 if all the VNFs of the NSPR are placed and 0 otherwise.
\end{itemize}
The values assumed by the above  variables are controlled by the problem constraints identified in  \cite{cnsm_2020} and \cite{HA_DRL_TNSM}. 

Optimally solving this problem using ILP based techniques would require a proper definition of the values for the three objective function coefficients and also affording for the long convergence times of ILP on hard problems. We propose instead in Section \ref{sec:HA-DRL} a faster and more scalable solution based on DRL controlled by a heuristic method.

\section{Heuristically Assisted DRL for Network Slice Placement \label{sec:HA-DRL}}

We transform the Network Slice Placement Optimization problem described in Section \ref{sec:pb} in a Partially Observable Markov Decision Process (POMDP) in which: i) the placement of one NSPR is calculated iteratively (i.e., VNFs are placed sequentially, one by one, starting from the first VNF, along with its associated VLs), ii) the state contains the features of the PSN (i.e., the CPU, RAM, and bandwidth available capacities --- $cap^{cpu}$, $cap^{ram}$, $cap^{bw}$, respectively ---  associated with each node; a placement mask $x$ giving the number of VNFs of the current NSPR placed in each PSN node) and NSPR (i.e., the CPU, RAM and bandwidth requirements --- $req^{cpu}$, $req^{ram}$, $req^{bw}$, respectively --- associated with each VNF of the NSPR and a counter $m_v$ of missing VNFs to place at each iteration), iii) the action calculated by the policy is a valid placement (i.e., a placement respecting the problem constraints), and iv) the reward corresponds to an appropriate function that measures how good is the computed action with respect to the optimization objectives described in Section \ref{sec:pb}. We apply the proposed HA-DRL concept to this problem.

\begin{figure}[hbtp] 
\centering
\includegraphics[width=\linewidth]{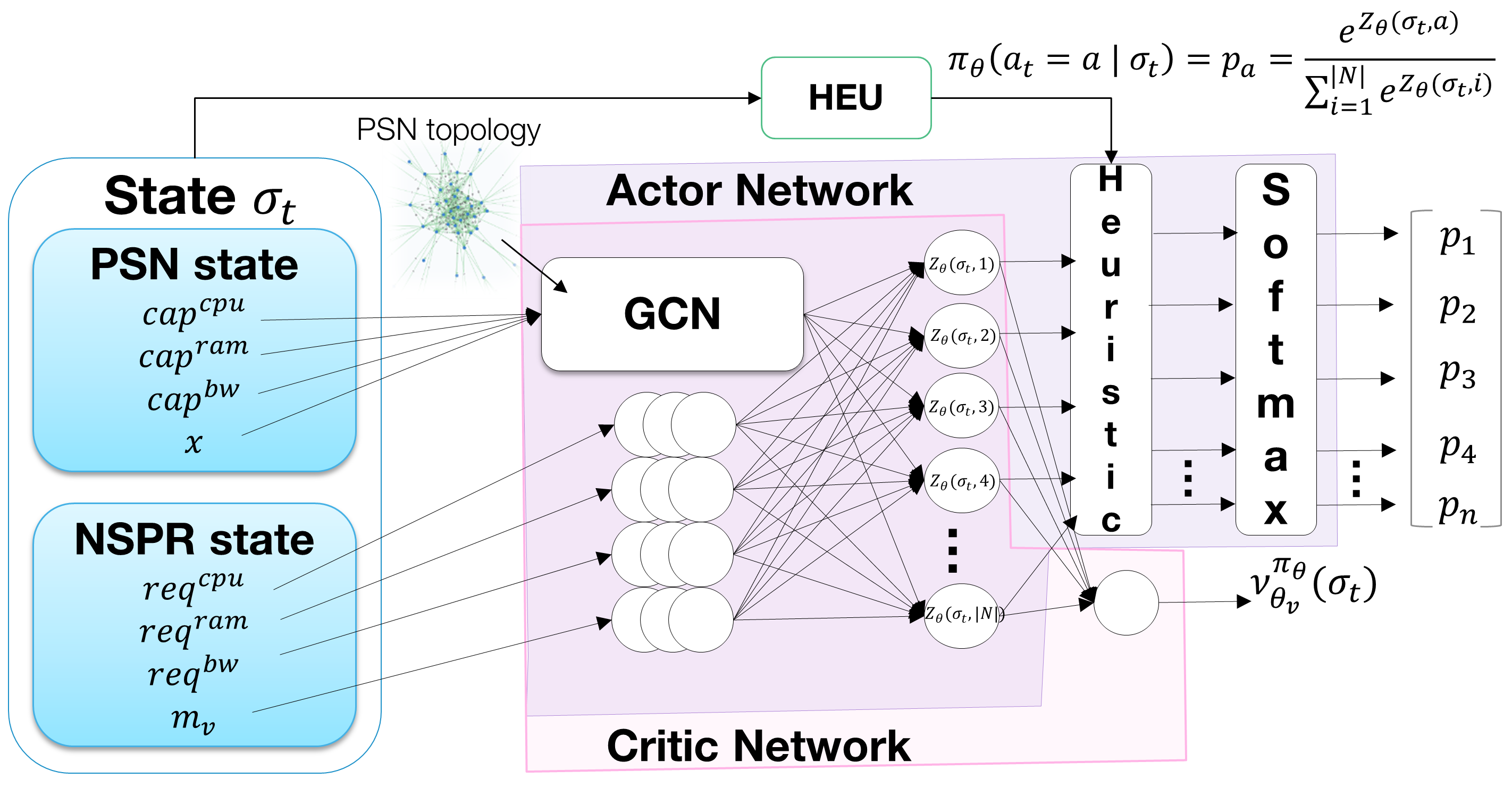}
\caption{Structure of the proposed Heuristically Assisted DRL algorithm} \label{fig::ha_advantage_actor_critic_architecture}
\end{figure}

Fig.~\ref{fig::ha_advantage_actor_critic_architecture} presents the structure of the proposed HA-DRL algorithm. It is an extension of the Asynchronous Advantage Actor Critic (A3C) algorithm introduced in \cite{p1},  referred to in the present paper   as DRL, for the VNE problem. DRL uses two Deep Neural Networks (DNNs) to learn the optimal policy $\pi_{\theta}$ and the optimal state-value $\nu^{\pi_{\theta}}_{\theta_v}$ function called  Actor and Critic Networks, respectively. We propose to modify the Actor Network by adding a Heuristic Function layer that enhances the exploration process and accelerates the convergence of the algorithm by influencing the policy choice of actions. This layer benefits from external information provided by an efficient heuristic for slice placement we proposed in \cite{cnsm_2020}, referred  as HEU. A detailed description of the proposed HA-DRL algorithm can be found in \cite{HA_DRL_TNSM}.

\section{Implementation \& Evaluation Results \label{sec:evaluation}}

We developed a simulator in Python to implement and evaluate the proposed DRL and HA-DRL algorithms. We used the PyTorch framework to implement the DNNs. We consider an implementation of the HEU algorithm \cite{cnsm_2020} in Julia as a benchmark in the performance evaluation experiments. Experiments were run in a 2x6 cores @2.95Ghz 96GB machine. We consider NSPRs to have the Enhanced Mobile Broadband (eMBB) settings described in \cite{cnsm_2020} with 5 to 20 VNFs. We consider a PSN that could reflect the infrastructure of an operator such as Orange \cite{cnsm_2020} with 126 to 1008 placement nodes. DRL and HA-DRL agents are trained for 24 hours  with learning rates for the Actor and Critic networks set to $\alpha = 10^{-4}$ and $\alpha' = 2.5. 10^{-3}$ respectively.

We consider four versions of HA-DRL algorithms each with a different value for the $\beta$ parameter of the Heuristic Function (HA-DRL, $\beta = 0.1$; HA-DRL, $\beta = 0.5$; HA-DRL, $\beta = 1.0$; HA-DRL, $\beta = 2.0$). The value of $\beta$ controls how much the HEU method influences the policy. The higher the value of $\beta$ is, the more the HA-DRL algorithm is most likely to select the action computed by the HEU method, see \cite{HA_DRL_TNSM}.

\subsection{Acceptance Ratio Evaluation \label{sec:ar_evaluation}}

\begin{figure}[ht] 
\centering
\begin{subfloat}[Under-loaded scenario]
 {\includegraphics[width=.55\linewidth]{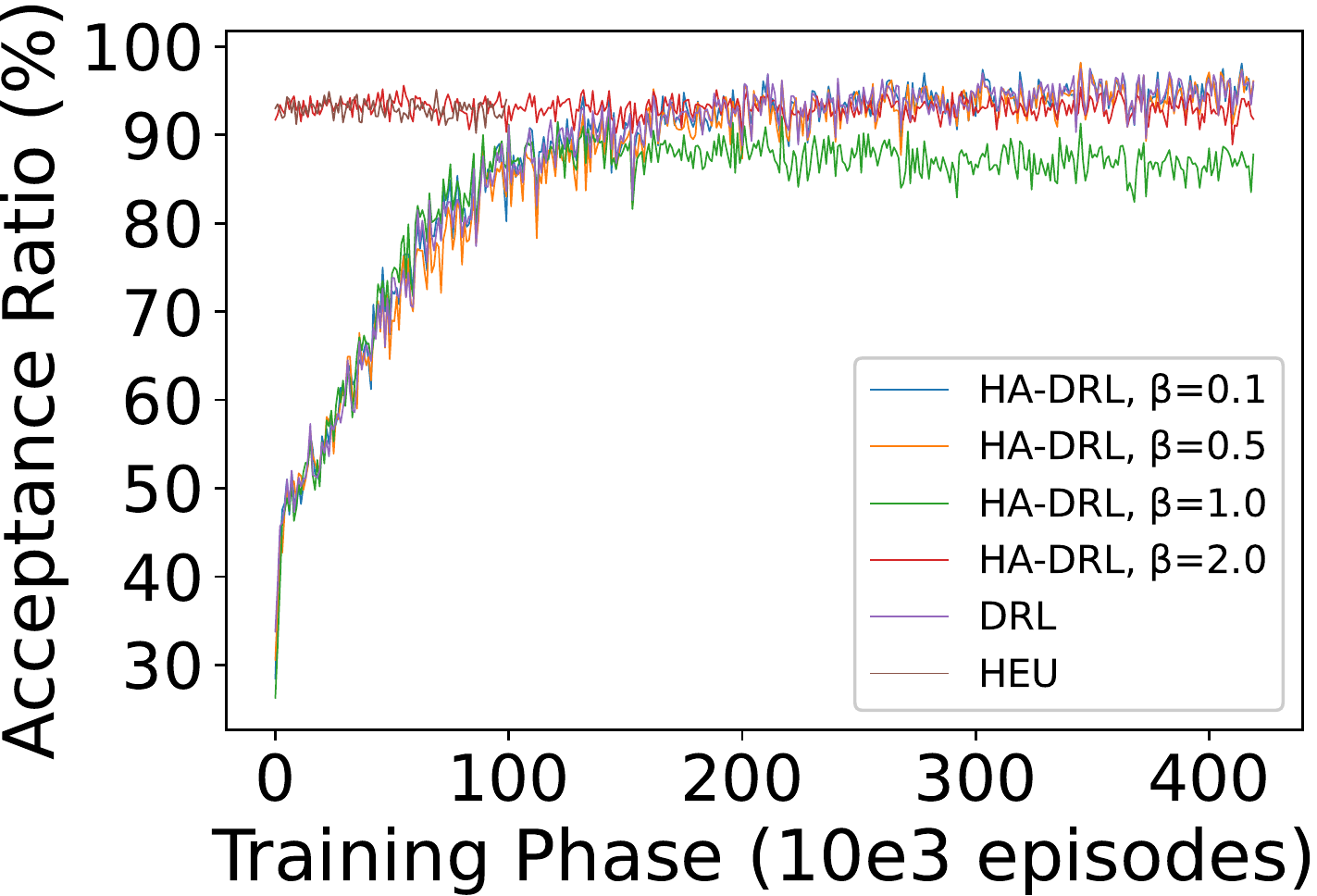}\label{fig:acc_ratio_0.5}}
\end{subfloat}
\begin{subfloat}[Critical load scenario]
 {\includegraphics[width=.55\linewidth]{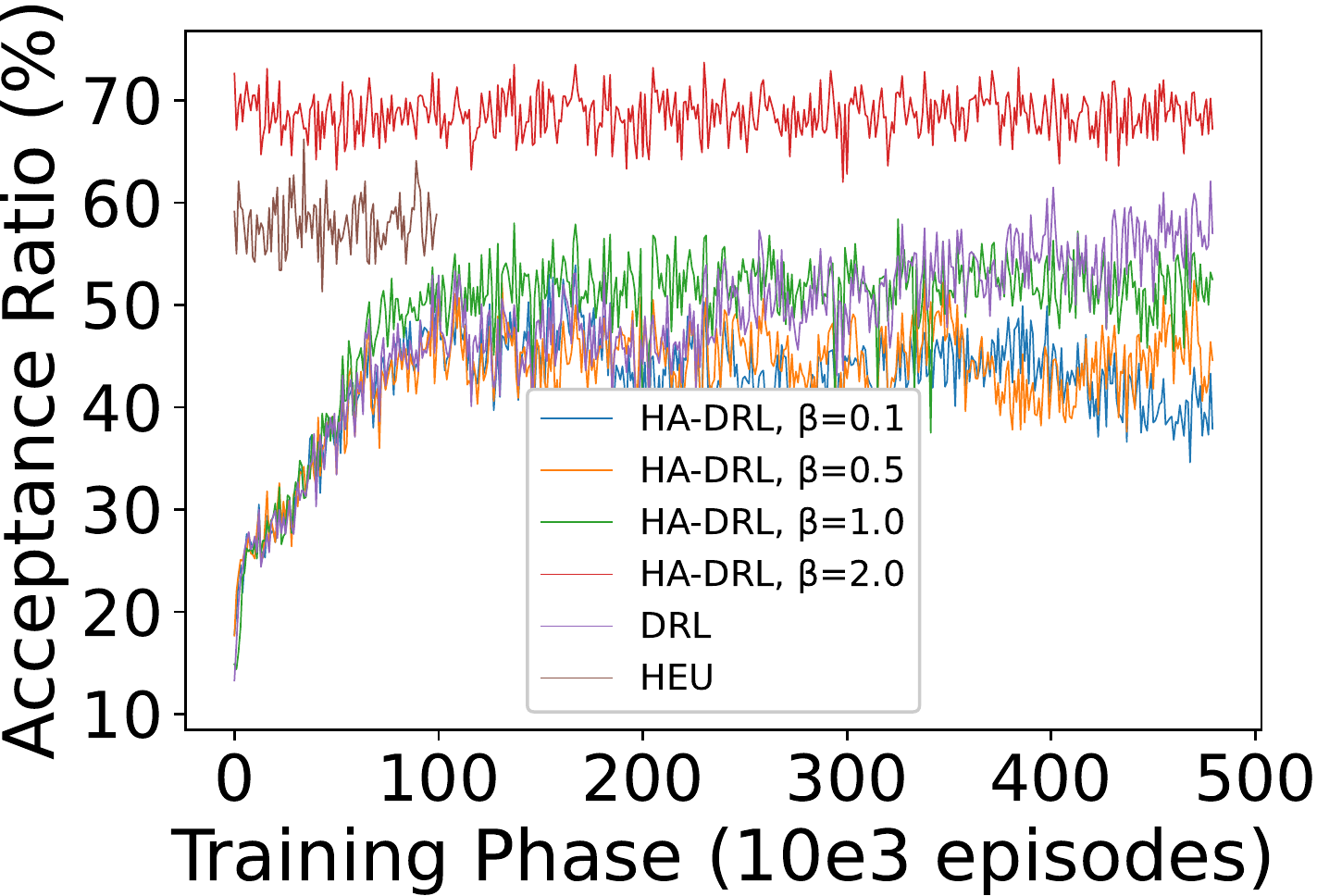}\label{fig:acc_ratio_1.0}}
\end{subfloat}
\caption{Acceptance ratio evaluation results.}
\label{fig:ar_vs_tr_phase}
\end{figure}

Fig.~\ref{fig:ar_vs_tr_phase} and Tables \ref{tab:ars_0.5}, \ref{tab:ars_1.0} present the Acceptance Ratios obtained with the HA-DRL, DRL and HEU algorithms in under-loaded (50\%) and critical (100 \%) regimes. HA-DRL with parameter $\beta = 2.0$ shows the best performance with fast convergence in both scenarios. This happens because when setting $\beta = 2.0$ the Heuristic Function calculated on the basis of the HEU algorithm has strong influence on the actions chosen by the algorithm. Since the HEU algorithm often indicates a good action, this helps the algorithm to become stable more quickly.

Fig.~\ref{fig:acc_ratio_0.5} and Table \ref{tab:ars_0.5} reveal that DRL and HA-DRL algorithms with $\beta \in \{0.1, 0.5,1.0\}$ converge within an interval of 200 to 300 training phases in the under-loaded scenario and that all algorithms except HA-DRL with $\beta = 1.0$ have an Acceptance Ratio higher than 94\% in the last training phase.

Fig.~\ref{fig:acc_ratio_1.0} and Table \ref{tab:ars_1.0} show that in the Critical load scenario, HA-DRL with $\beta=2.0$ performs significantly better than the other algorithms at the end of the training phase. HA-DRL with $\beta=2.0$ accepts 67.2~\% of the NSPR arrivals in the last training phase, 8.34~\% more than HEU, 10.21~\% more than DRL, 14.7~\% more than HA-DRL with  $\beta = 1.0$, 22.6\% more than HA-DRL with $\beta = 0.5$, 29.3\% more than HA-DRL with $\beta = 0.1$. Fig.~\ref{fig:acc_ratio_1.0} also shows that the performance of algorithms DRL and HA-DRL with $\beta \in \{0.1, 0.5,1.0\}$ is still not stable at the end of the training process in the Critical load scenario.

\begin{table}[t]
\begin{threeparttable}
\centering
\caption{Acceptance Ratio at different Training Phases, Under-loaded Scenario}
\label{tab:ars_0.5}
\begin{tabularx}{\linewidth}{@{}cLLLLL@{}}
\toprule
\multirow{2}{*}{\textbf{Algorithm}} & \multicolumn{5}{c}{\textbf{Acceptance Ratio at different Training Phases (\%)}} \\ \cmidrule(l){2-6} 
                   &25&100&200&300&400    \\ \midrule
HA-DRL,$\beta$=0.1&57.90& 80.20&92.70&93.00&96.20\\
HA-DRL,$\beta$=0.5&58.50& 83.00&90.20&94.80&96.30\\
HA-DRL,$\beta$=1.0&58.80& 86.20&86.80&85.50&85.80\\
HA-DRL,$\beta$=2.0&93.50& 90.30&93.10&92.20&94.80\\
DRL                &57.10& 83.60&91.20&94.80& 95.70\\
HEU                &93.50& 94.00&94.00*&94.00 &94.00*\\ \bottomrule
\end{tabularx}
\begin{tablenotes}
    \footnotesize
     \item $^{*}$ Performance of the HEU algorithm in  the steady state.
    \end{tablenotes}
\end{threeparttable}
\end{table}

\begin{table}[t]
\caption{Acceptance Ratio at different Training Phases, Critical Load Scenario}
\label{tab:ars_1.0}
\begin{tabularx}{\linewidth}{@{}cLLLLLL@{}}
\toprule
\multirow{2}{*}{\textbf{Algorithm}} & \multicolumn{6}{c}{\textbf{Acceptance Ratios at different Training Phases (\%)}} \\ \cmidrule(l){2-7} 
                                    &25&100&200&300&400&480\\ \midrule
HA-DRL,$\beta$=0.1&30.10&49.70&45.30&45.0&41.00&37.90\\
HA-DRL,$\beta$=0.5&30.80&45.90&50.80&44.5&38.90&44.60\\
HA-DRL,$\beta$=1.0&27.40& 52.00&55.50&55.60&49.00&52.50\\
HA-DRL,$\beta$=2.0&67.60&67.60&70.80&69.60&66.10&67.2\\
DRL&29.30&49.10&46.60&50.10&53.40&56.99\\
HEU&60.70&58.86&58.86*&58.86*&58.86*&58.86*\\ \bottomrule
\end{tabularx}

\end{table}

\begin{figure}[ht] 
\centering
\begin{subfloat}[]
 {\includegraphics[width=0.58\linewidth]{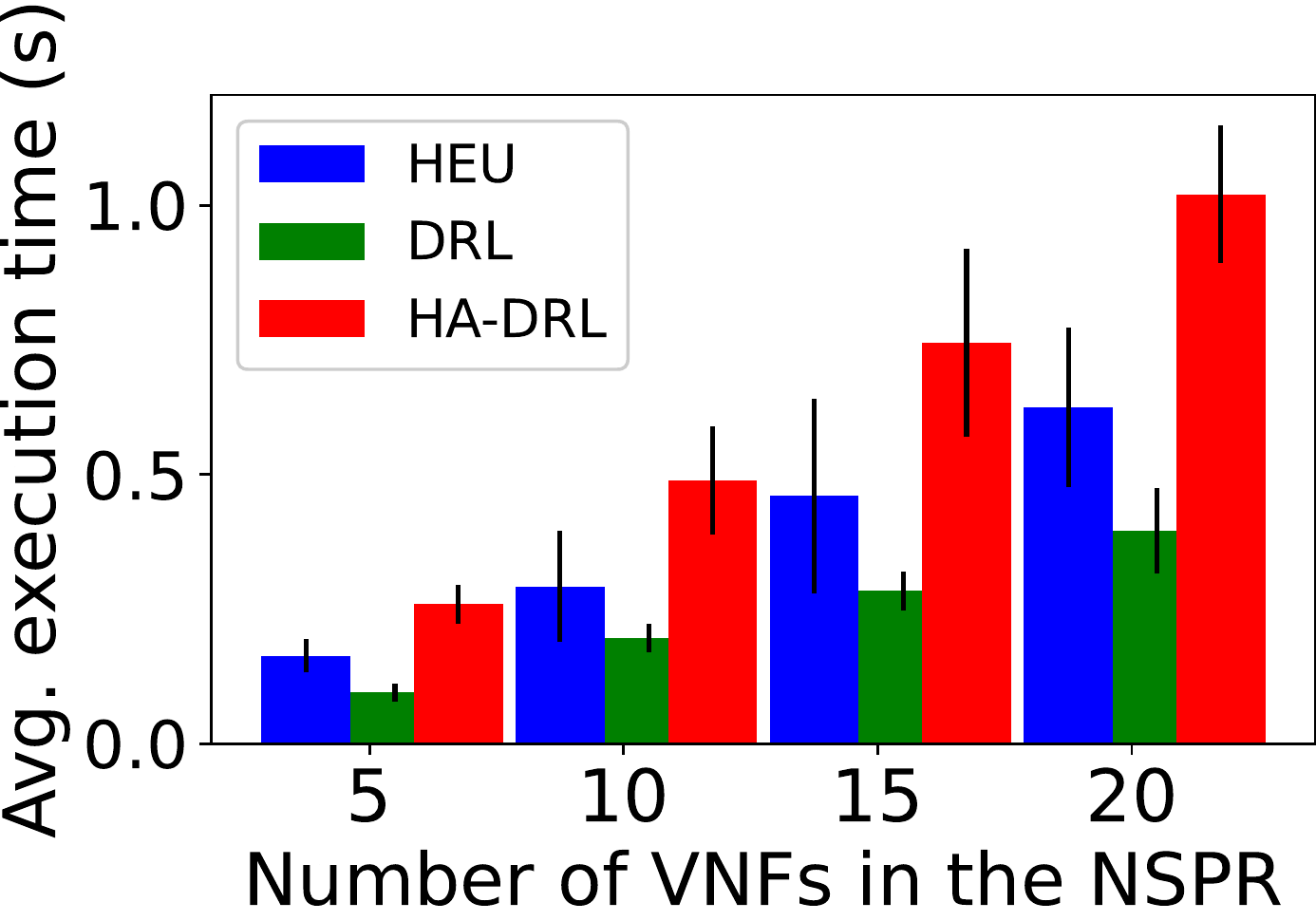}\label{fig:avg_exec_time_vs_nspr_size}}
\end{subfloat}
\begin{subfloat}[]
 {\includegraphics[width=0.58\linewidth]{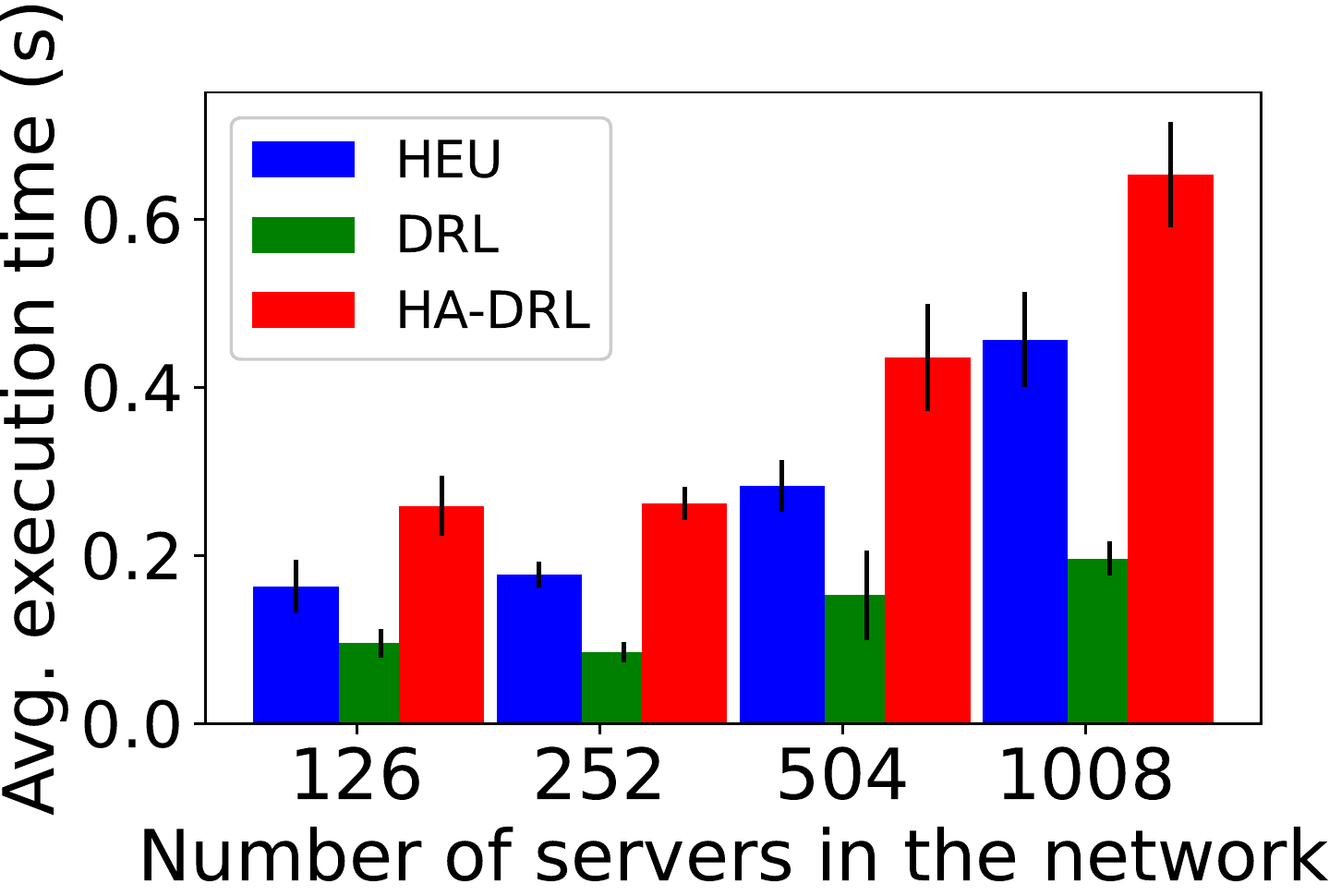}\label{fig:avg_exec_time_vs_nodes}}
\end{subfloat}
\caption{Average execution time evaluation.}
\label{fig:avg_exec_time}
\end{figure}

\subsection{Execution Time Evaluation \label{sec:ex_time_evaluation}}

Fig.~\ref{fig:avg_exec_time_vs_nspr_size} and \ref{fig:avg_exec_time_vs_nodes} present the average execution time of the HEU, DRL and HA-DRL algorithms as a function of the number of VNFs in the NSPR and the number of servers in the PSN, respectively. The evaluation results show that the average execution times increase faster for heuristics than for a pure DRL approach. 

However, both HEU and DRL strategies have low execution times (less than 0.6s in the largest scenarios). HA-DRL has an average execution time approximately equal to the sum of HEU and DRL  execution times. 

Since DRL and HEU have small execution times, the average execution times of HA-DRL are also small.

\section{Conclusion \label{sec:conclusion}}

We have briefly described the Heuristically Assisted DRL (HA-DRL) approach to Network Slice Placement Optimization introduced in  \cite{HA_DRL_TNSM}. The main contributions of the proposed solution are: i) enhanced scalability compared to classical ILP and heuristic approaches, ii) adoption of multiple optimization criteria in a simple way, iii) accelerated learning by exploiting external information provided by an efficient heuristic. Evaluation results show that the proposed HA-DRL approach yields good placement solutions in nearly real time, converges significantly faster than pure DRL approaches, and yields better performance in terms of acceptance ratio than state-of-the-art heuristics and DRL algorithms when used alone. As part of our future work, we plan to explore distribution and  parallel computing techniques to solve the considered  multi-objective optimization problem using a multi-agent or federated learning to assess scalability of network slices.

\section*{Acknowledgment}

This work has been performed in the framework of 5GPPP MON-B5G project (www.monb5g.eu).

%

\bibliographystyle{IEEEtran}
\bibliography{IEEEabrv,my_bib}

\begin{thebibliography}{1}
\providecommand{\url}[1]{#1}
\csname url@samestyle\endcsname
\providecommand{\newblock}{\relax}
\providecommand{\bibinfo}[2]{#2}
\providecommand{\BIBentrySTDinterwordspacing}{\spaceskip=0pt\relax}
\providecommand{\BIBentryALTinterwordstretchfactor}{4}
\providecommand{\BIBentryALTinterwordspacing}{\spaceskip=\fontdimen2\font plus
\BIBentryALTinterwordstretchfactor\fontdimen3\font minus
  \fontdimen4\font\relax}
\providecommand{\BIBforeignlanguage}[2]{{%
\expandafter\ifx\csname l@#1\endcsname\relax
\typeout{** WARNING: IEEEtran.bst: No hyphenation pattern has been}%
\typeout{** loaded for the language `#1'. Using the pattern for}%
\typeout{** the default language instead.}%
\else
\language=\csname l@#1\endcsname
\fi
#2}}
\providecommand{\BIBdecl}{\relax}
\BIBdecl

\bibitem{survey_vfnp}
A.~Laghrissi and T.~Taleb, ``A survey on the placement of virtual resources and
  virtual network functions,'' \emph{{IEEE} Commun. Surveys Tuts.}, vol.~21,
  no.~2, pp. 1409--1434, 2nd. Quart., 2019.

\bibitem{vne_np_hardness}
E.~Amaldi, S.~Coniglio, A.~M. Koster, and M.~Tieves, ``On the computational
  complexity of the virtual network embedding problem,'' \emph{Electron. Notes
  Discrete Math.}, vol.~52, pp. 213--220, Jun. 2016.

\bibitem{cnsm_2020}
J.~J. {Alves Esteves}, A.~{Boubendir}, F.~{Guillemin}, and P.~{Sens},
  ``Heuristic for edge-enabled network slicing optimization using the “power
  of two choices”,'' in \emph{Proc. 2020 IEEE 16th Int. Conf. Netw. Service
  Manag. (CNSM)}, 2020, pp. 1--9.

\bibitem{p1}
Z.~{Yan}, J.~{Ge}, Y.~{Wu}, L.~{Li}, and T.~{Li}, ``Automatic virtual network
  embedding: A deep reinforcement learning approach with graph convolutional
  networks,'' \emph{{IEEE} J. Sel. Areas Commun.}, vol.~38, no.~6, pp.
  1040--1057, Jun. 2020.

\bibitem{HA_DRL_TNSM}
J.~J. {Alves Esteves}, A.~{Boubendir}, F.~{Guillemin}, and P.~{Sens}, ``A
  heuristically assisted deep reinforcement learning approach for network slice
  placement,'' \emph{arXiv preprint arXiv:2105.06741}, 2021.

\end{thebibliography}


\end{document}